\documentclass{article}
\usepackage{spconf,amsmath,graphicx}

\usepackage{enumitem}
\setlist{nosep, leftmargin=14pt}

\usepackage{mwe} % to get dummy images
\usepackage{booktabs}
\usepackage{tikz}
\usepackage{afterpage}
\usepackage{amsfonts}
\usepackage{multirow}
\usepackage{url}
\usepackage{hyperref}

\title{DINOv2 based Self Supervised Learning \\ For Few Shot Medical Image Segmentation}
\name{Lev Ayzenberg$^1$, Raja Giryes$^1$ and Hayit Greenspan$^{1,2}$}
\address{$^1$Faculty of Engineering, Tel Aviv University \quad $^2$ Icahn school of Medicine, Mount Sinai,NY}
\begin{document}
%\ninept
%
\maketitle
\begin{abstract}
Deep learning models have emerged as the cornerstone of medical image segmentation, but their efficacy hinges on the availability of extensive manually labeled datasets and their adaptability to unforeseen categories remains a challenge. Few-shot segmentation (FSS) offers a promising solution by endowing models with the capacity to learn novel classes from limited labeled examples. A leading method for FSS is ALPNet, which compares features between the query image and the few available support segmented images. A key question about using ALPNet is how to design its features. 
In this work \footnote{The code will be made available at \url{https://github.com/levayz/DINOv2-based-Self-Supervised-Learning.git}}
, we delve into the potential of using features from DINOv2, which is a foundational self-supervised learning model in computer vision. Leveraging the strengths of ALPNet and harnessing the feature extraction capabilities of DINOv2, we present a novel approach to few-shot segmentation that not only enhances performance but also paves the way for more robust and adaptable medical image analysis. 
\end{abstract}
\section{Introduction}
\label{sec:intro}

Deep learning models have firmly established themselves as the primary approach to medical image segmentation. Yet, conventional deployment of deep learning for medical image segmentation often demands a substantial amount of manually annotated data for effective training, which can be a costly and labor-intensive endeavor. Furthermore, these models face challenges when confronted with previously unseen categories, necessitating further training and adaptation.

To address these limitations, few-shot segmentation (FSS) emerged as a potential solution \cite{Wang2020Generalizing}. Few-shot segmentation trains the model to learn and generalize from a limited number of labeled examples, thereby alleviating the need for extensive, manually annotated datasets.

Among the various different FSS techniques, Prototypical Networks (PN) \cite{snell2017prototypical} is a popular choice for few shot learning. These networks utilize prototypes, which encapsulate the essential features of semantic classes, enabling similarity-based predictions. One such approach is ALPNet \cite{alpnet}, which achieves the current state-of-the-art (SOTA) in FSS in medical applications. Their main innovation is the introduction of the Adaptive Local Prototypes Pooling module (ALP), which improves  capturing fine-grained details in medical images. 

An important factor in the performance of PN is the features being used. One strategy is to use features from pre-trained deep networks for other tasks. In this work, we consider the case of using self-supervised learning, where a neural network is trained to produce good representations for given data without having any labels for them. Specifically, we employ DINOv2 \cite{dinov2} features for our task. DINOv2 is a foundational model in self-supervised learning that is based on a transformer architecture and provides an improved representation compared to prior models. By harnessing DINOv2 capabilities, we aim to improve FSS performance. 

We explore various options for using DINOv2. Specifically, we show that incorporating DINOv2 as an encoder within ALPNet combined with connected component analysis (CCA) and test time training (TTT) leads to improved performance in various medical segmentation datasets.
%In addition, we show that adding connected component analysis (CCA) to FSS further improves the results. 

\section{Related Work}
\label{sec:related_work}

\noindent {\bf Few Shot Segmentation.}
In standard medical image segmentation using deep learning models, neural networks are trained in a fully supervised way to predict a per-pixel label for the input images. Given a new segmentation task, this usually entails starting from scratch (perhaps with a pre-trained network for classification), requiring significant design and tuning, as well as access to substantial annotated datasets. As a more practical solution, few-shot segmentation (FSS) offers an efficient cost-effective approach that enables models to excel with limited annotated data.

FSS refers to training a model that can segment new classes by introducing additional
prior knowledge in the form of a small `support' annotated set.
Prototypical Networks (PN) is a popular choice for addressing few-shot learning tasks. They focus on exploiting representation prototypes of semantic classes extracted from the support. These prototypes are utilized to make similarity-based predictions.
A recent PN approach, ALPNet \cite{alpnet} introduces the Adaptive Local Prototypes Pooling (ALP) module. ALP is a computation module responsible for deriving both local and class-level prototypes. It enhances the model's ability to capture fine-grained details.

Another work, which relies on ALPNet, proposed a cross-reference transformer, aiming to enhance the similar parts of support features and query features in high-dimensional channels  \cite{cross_reference_transformer}.
CRAP-Net \cite{crapnet}, which also relies on ALPNet, introduced an attention mechanism to enhance the relationship between support and query pixels to preserve the spatial correlation between image features. It smoothly incorporates this mechanism into the conventional prototype network.

\noindent {\bf Self-Supervised Learning} methods learn visual features from unlabeled data. The unlabeled data is used to automatically generate pseudo labels for a pretext task. In the course of training to solve the pretext task, the network learns visual features that can be transferred to solving other tasks with little to no labeled data \cite{gui2023survey}. In this work, we employ
DINOv2 \cite{dinov2}, which is a recent self-supervised learning model. Its architecture is based on the vision transformer (ViT) model \cite{vit}. DINOv2 learns a representation for natural images that can then be adapted to various computer vision tasks including object detection, segmentation and depth estimation. The trained models that generate these representations are referred to as the DINOv2 encoder.

\begin{figure}[t]
\centering
\includegraphics[width=\columnwidth]{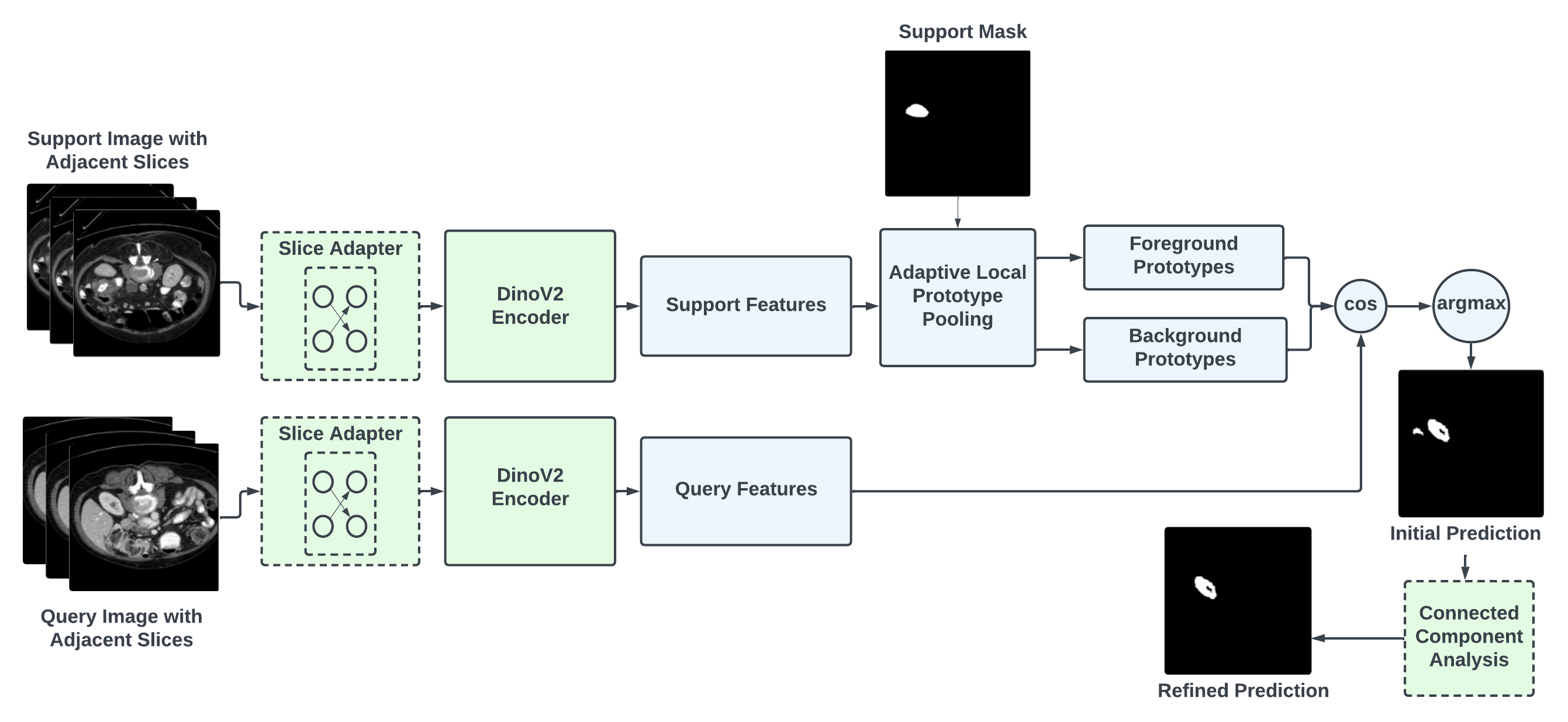}
\vspace{-0.35in}
\caption{Proposed Framework: Blue - original ALPNet architecture, Green - added or replaced components, Dotted line - Optional component}  % Add your desired caption here.
\vspace{-0.15in}
\label{fig:alpnet_arch}
\end{figure}

\section{Method}
\label{sec:method}

\noindent \textbf{Problem Formulation.}
The objective of few-shot segmentation is to train a function, denoted as $f(\mathbf{x}^q, S)$, capable of predicting a binary mask for an unseen class when provided with a query image, $\mathbf{x}^q$, and the support set, $S$.
The support image set, denoted as $S$, comprises pairs ${(\mathbf{x}^s_i(c), \mathbf{y}^s_i(c))_{i=1}^k, c\in C_{test}}$. Here, $\mathbf{x}^s_i$ represents the $i$-th image in the support image set, and $\mathbf{y}^s_i(c)$ represents the segmentation mask of the $i$-th support image corresponding to class $c$.
The dataset is split into two parts: the training dataset, $D_{train}$, and the testing dataset, $D_{test}$. These datasets consist of image-binary mask pairs, with $D_{train}$ annotated by $C_{train}$ and $D_{test}$ annotated by $C_{test}$.
There are no overlapping classes between these two sets, meaning that $C_{train} \cap C_{test} = \emptyset$.

During the training of our few-shot networks, the model processes input data in the form of $\langle S, \mathbf{x}^q \rangle$ pairs, with $S$ being a subset of $D_{train}$ ($S \subset D_{train}$). Note that $(\mathbf{x}^q, \mathbf{y}^q(c)) \notin S$, and information from $\mathbf{y}^q(c)$ is solely used for training purposes.
Each instance of such a pair is referred to as an "episode," with each episode being randomly selected from the $D_{train}$ dataset. The support set $S$ comprises a total of $k$ image-binary mask pairs for the semantic class $c$, and there are $n$ classes within $C_{test}$. This structure characterizes it as an "n-way k-shot segmentation sub-problem" for each episode.

\noindent \textbf{Network Architecture.}
As seen in Figure~\ref{fig:alpnet_arch}, the network is composed of an encoder, and an adaptive local prototype pooling module (ALP) \cite{alpnet} for extracting prototypes.
The support images go through the encoder, and the encoder provides the ALP module with feature maps of the support images.
The ALP module takes the provided feature maps along with their masks and outputs global and local prototypes, both for the foreground class and the background.
The ALP module performs average pooling with a pooling window of size $(L_H, L_W)$, for each feature map $f_{\theta}(\mathbf{x}^s_i) \in \mathbb{R}^{D\times H\times W}$ of the support image $\mathbf{x}^s_i$, where $(H,W)$ is the spatial size of the feature map and $D$ is the number of channels.
The prototype of example $i$ at location $(m,n)$ is calculated using
\begin{equation}
\label{eq:proto_calc}
    p_{i,m,n}(c) = \frac{1}{L_H L_W} \sum_{h} \sum_{w} f_{\theta}(\mathbf{x}^s_i(c)(h,w),
\end{equation}
where $mL_H \leq h < (m+1) L_H, \ nL_W \leq w < (n+1)L_W$.\\

In addition to local prototypes, a class-level prototype $p_i^g(c)$ is calculated, using equation~\ref{eq:class_level_proto}, where $\mathbf{y}_i^s(c)$ is the binary mask of class $c$ in $\mathbf{x}_i^s(c)$. The purpose of the class-level prototype is to ensure at least a single prototype is generated for objects smaller than the pooling window.

\begin{equation}
    p_i^g(c) = \frac{\sum\limits_{h,w}\mathbf{y}_i^s(c)(h,w)f_{\theta}(\mathbf{x}_i^s(c))(h,w)}{\sum\limits_{h,w}\mathbf{y}_i^s(c)(h,w)}
    \label{eq:class_level_proto}
\end{equation}
Then the prototypes are grouped together to a set $P = \{p_l(c)\}$. This includes the global and local prototypes.

% To predict the class of a pixel at location $(h,w)$, $Y_{q}(h,w)$, we compute the cosine similarity between the query features and the class prototypes, and decide the class based on the higher similarity. We use the follwing formulas:

A local similarity map for each class $c^j$ and each prototype is computed using the following formula:
\begin{equation}
    S_{l}(c^j)(h, w) = 20\cdot p_{l}(c^j)\odot f_{\theta}(\mathbf{x}^q)(h,w)
    \label{eq:local_sim_map}
\end{equation}
Where $\odot$ denotes cosine similarity, $l$ is the index for the prototype and $\mathbf{x}^q$ is the query image.\\
Then the local similarity maps are fused for each class separately into class-wise similarities $S'(c^j)$, using
\begin{equation}
S'(c^j)(h,w) = \underset{l}{\sum} S_l(c^j)(h,w) \cdot \text{softmax}[S_l(c^j)(h,w)]
\label{eq:global_prob}
\end{equation}
To obtain the final prediction, class-wise similarities are normalized into probabilities using:
\begin{equation}
\label{equ: pred_prob}
\mathbf{\hat{y}}^q(h,w) = \underset{j}{\text{softmax}}[S'(c^j)(h,w) ]
\end{equation}

\noindent \textbf{Training.}
The training procedure is the same as in \cite{alpnet}. We emulate real-life scenarios by structuring episodes. In each episode, we focus on a single slice. Initially, we generate superpixels for all available slices, using Felzenszwalb \cite{felzenszwalb2004efficient}, a preprocessing step performed before training. 
%From these superpixels, we randomly select one to act as a label. 
% Subsequently, we subject the original slice and its pseudo label to two geometric transformations, resulting in two variations: one designated as the support set and the other as the query set. The primary objective of this setup is to perform superpixel segmentation on the query slice.
At each episode an image $\mathbf{x}_i$ is chosen together with a random superpixel $\mathbf{y}_i^r(c^p)$ to form the support set $S_i={(\mathbf{x_i}, \mathbf{y}_i^r(c^p))}$. $c^p$ denotes the superpixel class, and $r$ the index of the random superpixel. The query set is formed by augmenting the chosen image $\mathbf{x}_i$ i.e $Q_i={(\mathcal{T}_g(\mathcal{T}_i(\mathbf{x}_i}))$, where $\mathcal{T}_g$ and $\mathcal{T}_i$ are geometric and intensity transforms respectively.
We employ equation~\ref{equ:loss_seg} as the cross-entropy loss.
\begin{eqnarray}
\label{equ:loss_seg}
&& \hspace{-0.3in} \mathcal{L}^i_{\text{seg}}(\theta ; S_i, Q_i) = \\
\nonumber && \hspace{-0.3in}
- \frac{1}{HW} \sum_{h=1}^H \sum_{w=1}^W \sum_{j \in \{0,p\}} \mathcal{T}_g(\mathbf{y}_i^r(c^j))(h,w) \log( \hat{\mathbf{y}}_i^r(c^j)(h,w) ),
\end{eqnarray}
where $\hat{\mathbf{y}}_i^r(c^p)$ is the prediction of the pseudolabel, $c^0$ is the background class and $\theta$ - the models parameters.

Also, as in \cite{alpnet}, we incorporate the \textit{prototype alignment regularization} \cite{wang2020panet}, where the roles of the support label and the prediction are reversed. The prediction assumes the role of the support label, and our aim is to segment the original superpixel accordingly.
The regularization loss is
\begin{eqnarray}
\label{equ:loss_reg}
&& \hspace{-0.3in} \mathcal{L}^i_{\text{reg}}(\theta ; \mathcal{S}'_i, \mathcal{S}_i) = \\ \nonumber && \hspace{-0.3in}
- \frac{1}{HW} \sum_{h=1}^H \sum_{w=1}^W \sum_{j \in \{0,p\} } \mathbf{y}^r_i(c^j)(h,w) \log( \bar{\mathbf{y}}^r_i(c^j)(h,w) ),
\end{eqnarray}
where $\bar{\mathbf{y}}^r_i(c^j)$ is the prediction of the superpixel label $\mathbf{y}^r_i(c^p)$.

\noindent \textbf{Encoder.}
The Encoder has several configurations: (i) An encoder that receives a single ct slice (encoded as an RGB image by repeating the slice across the RGB channels);
(ii) an encoder that receives 3 consecutive ct slices by encoding each slice as a separate channel of an RGB image; and
(iii) an adapter that serves to transform the 3 slices to a single image which then goes into the encoder. We use a simple linear layer for that.
The available encoders are: (i) default encoder used in ALPNet (deeplabv2 \cite{chen2017deeplab}); and (ii) DINOv2 \cite{dinov2} encoder.

\noindent \textbf{Inference.}
Just like in training, a single image and its mask are given (from the training set) to act as the support set. The model segments each slice for each scan from the test set using the support set.
After the initial segmentation, we employ Connected Component Analysis (CCA), on the results to choose the most confident component using equation.
The confidence of a connected component is derived from:
\begin{equation}
\text{Confidence} = \frac{\sum_{i} p_{i} \cdot \mathbf{\hat{y}}_i}{\sum_{i} \mathbf{\hat{y}}_i},
\label{eq:conf}
\end{equation}
where $p_i$ is the probability that pixel $i$ belongs to the foreground class, $\mathbf{\hat{y}}_i$ is the pixel $i$'s predicted label.

\noindent \textbf{Slice Adapter.} For the slice adapter, we select three consecutive slices from the input scan (denoted as $z_{i-1}$, $z_{i}$, and $z_{i+1}$), where $z_{i}$ is the slice to be segmented. The slices are then fed into the slice adapter. For the slice adapter we use a convolutional layer with $k=1$ and 3 output channels. The output of the slice adapter is then fed into the encoder.

\noindent \textbf{Test Time Training (TTT).}
To enhance our results, we implement self-supervised test-time training. In TTT, at inference time, we segment the test set as described above and save the labels. Subsequently, we iterate over the test set, augmenting each slice, $\mathbf{x}_i$, along with its predicted segmentation label, $\hat{\mathbf{y}_i}$ , through geometric and intensity augmentations, $\Tilde{\mathbf{x}_i}=\mathcal{T}_g(\mathcal{T}_i(\mathbf{x}_i))$ , $\Tilde{\mathbf{y}_i}=\mathcal{T}_g(\hat{\mathbf{y}_i})$. We then train the model to segment the augmented slice, $\Tilde{\mathbf{x}_i}$ using the augmented predicted label, $\Tilde{\mathbf{y}_i}$ as the ground truth. This method is similar to the regular training process described above with the difference that we replace the superpixels with our predicted labels.

\section{Experiments and Results}
\label{sec:experiments}
We employ two datasets for abdominal organ segmentation, each associated with a different modality (CT and MRI). These datasets include (i) Abd-CT: Derived from the MICCAI 2015 Multi-Atlas Abdomen Labeling challenge \cite{miccai2015}, containing 30 3D abdominal CT scans; and (ii) Abd-MRI: Sourced from the ISBI 2019 Combined Healthy Abdominal Organ Segmentation Challenge (Task 5) \cite{CHAOS}, comprising 20 3D T2-SPIR MRI scans.
Images are re-formated as 2D axial (Abd-CT and Abd-MRI) slices, and resized to 256 × 256 for training, and to 672 x 672 for testing.

%For evaluation the segmentation results, we employ the Dice score, which is commonly used in medical image segmentation tasks.
To evaluate 2D segmentation on 3D volumetric images, we follow the evaluation protocol established by \cite{squeeze_and_excite}

In a 3D image, when dealing with each specific class denoted as $c^{j}$, we divide the images that fall between the top and bottom slices containing this class into equal sections. These sections, in our experiments, are set to be $C=3$ in number. For each of these sections, we choose the middle slice from the corresponding section of the support scan as a reference point. Then, this reference slice is used to guide the segmentation of all the slices within the current section in the query scan. It's important to note that the support and query scans are obtained from different patients.
For the experiment setting we use a setting introduced in \cite{alpnet}. In this setting the testing class may not appear during training, meaning any slice that contains the testing class is discarded during training. This setting is referred to as ``Setting 2'' in \cite{alpnet}.
As done in other works \cite{alpnet,crapnet}, we divide the organs into two groups: (Spleen, Liver), (Right Kidney, Left Kidney)
In each experiment all slices containing the testing group will be removed from the training data.
In all our experiments, we use 1-way 1-shot learning.
We report the results of adding CCA, using Test Time Trainig (TTT) and employing a linear layer to act as a "slice adapter" in Table~\ref{table:mri} and Table~\ref{table:ct}, where compare to the current SOTA methods.
We also evaluated other options of using DINOv2 to perform segmentation as seen in Table~\ref{table:ablation ct}.

\noindent \textbf{Implementation details.}
We use the code provided by ALPNet \cite{alpnet} and the DINOv2-large encoder \cite{dinov2}, which has 300 million parameters.
For finetuning the DINOv2 encoder we employ Low Rank Adaptation (LoRA) \cite{hu2021lora}.

\begin{table}[t]
\centering
\resizebox{\linewidth}{!}{
\begin{tabular}{lcccccc}
\toprule
\multirow{2}{*}{Method} & \multicolumn{2}{c}{Abdominal-MRI Lower} & \multicolumn{2}{c}{Abdominal-MRI Upper} &  \\
& LK & RK & Spleen & Liver & \textbf{Mean} \\
\cmidrule(lr){1-1}\cmidrule(lr){2-3}\cmidrule(lr){4-5}\cmidrule(lr){6-6}
SSL-ALPNet \cite{alpnet} & 73.63 & 78.39 & 67.02 & 73.05 & 73.02 \\
CRAP-Net \cite{crapnet} & 74.66 & 82.77 & \textbf{73.82} & 70.82 & 73.82 \\
CRTPNet \cite{cross_reference_transformer} & 76.74 & 80.15 & 70.07 & 73.36 & 75.08 \\
\hline
SSL-DINOv2 large & 75.06 & 80.21 & 71.86 & 73.5 & 75.16 \\
SSL-DINOv2 large + CCA & \textbf{81.43} & \textbf{84.80} & 73.30 & 74.20 & \textbf{78.43} \\ 
SSL-DINOv2 slice adapter & 75.40 & 79.10 & 72.09 & 72.63 & 74.80 \\
SSL-DINOv2 slice adapter + CCA & 80.81 & 81.19 & 73.32 & 72.92 & 77.06 \\
%SSL-DINOv2 Pseudo Support Labels & \textbf{85.8} & \textbf{86.1} & 65 & 68.3 & 76.3 \\
SSL-DINOv2 TTT + CCA & 80.37 & 82.29 & 73.4 & \textbf{74.72} & 77.39 \\
%SSL-DINOv2 SAM labels & 76.62 & 80.6 &  &  &  \\
%DINOv2 finetuned on support & 53.40 & 49.16 & 53.25 & 65.72 & 55.38 \\
\bottomrule
\end{tabular}
}
\vspace{-0.15in}
\caption{MRI Results (in Dice score) on abdominal images}
\vspace{-0.05in}
\label{table:mri}
\end{table}

\begin{table}[t]
\centering
\resizebox{\linewidth}{!}{
\begin{tabular}{lcccccc}
\toprule
\multirow{2}{*}{Method} & \multicolumn{2}{c}{Abdominal-CT Lower} & \multicolumn{2}{c}{Abdominal-CT Upper} \\
& LK & RK & Spleen & Liver & \textbf{Mean} \\
\cmidrule(lr){1-1}\cmidrule(lr){2-3}\cmidrule(lr){4-5}\cmidrule(lr){6-6}
SSL-ALPNet \cite{alpnet} & 63.34 & 54.82 & 60.25 & 73.65 & 63.02 \\
CRAP-Net \cite{crapnet}& 70.91 & 67.33 & 70.17 & 70.45 & 69.72 \\
CRTPNet \cite{cross_reference_transformer} & 66.37 & 61.05 & 67.92 & 73.88 & 67.31 \\
\hline
SSL-DINOv2 large & 69.96 & 66.4 & 73 & 76.40 & 71.44 \\
SSL-DINOv2 large + CCA & 66.4 & 69.96 & 74.60 & \textbf{81.67} & 73.16 \\ 
SSL-DINOv2 slice adapter & 70.63 & 66.98 & 73.22 & 77.75 & 72.14 \\
SSL-DINOv2 slice adapter + CCA & \textbf{73} & 65.98 & 74.71 & 78.6 & 73.07 \\
%SSL-DINOv2 Psuedo Support Labels & 56.1 & 59.1 & 70.2 & 74.6 & 65 \\
SSL-DINOv2 TTT + CCA & 65.65 & \textbf{73.05} & \textbf{74.76} & 81.41 & \textbf{73.72} \\
%SSL-DINOv2 SAM labels & 61.93 & 59.53 &  &  &  \\
%DINOv2 finetuned on support & 42.53 & 49.27 & 46.89 & 56.34 & 48.76 \\
\bottomrule
\end{tabular}
}
\vspace{-0.15in}
\caption{CT Results (in Dice score) on abdominal images}
\vspace{-0.15in}
\label{table:ct}
\end{table}

\afterpage{%
\begin{figure}[htb]
    \begin{minipage}[b]{1.0\linewidth}
      \centering
      \centerline{\includegraphics[width=8.5cm]{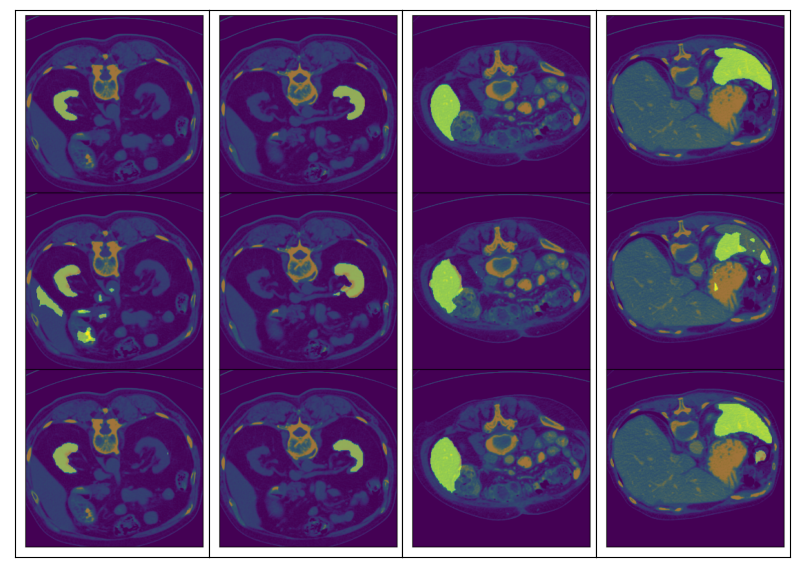}}
    %  \vspace{2.0cm}
    \end{minipage}
    \vspace{-0.35in}
    \caption{CT segmentation results for right kidney, left kidney, liver and spleen. The top row represents the ground-truth, the center row shows SSL-ALPNet with default encoder results and bottom row shows our results.}
    \label{fig:ct_seg}
    \vspace{-0.05in}
\end{figure}
}

\noindent \textbf{Comparison to other methods.} In order to assess the performance of our method, we compare the performance of our ALPNet and DINOv2 based model to SOTA models from recent years. Table \ref{table:mri} shows the results of our methods compared to the results of SOTA models under the same setting for MRI images, and Table~\ref{table:ct} shows
the results for CT images. Simply by replacing the encoder of SSL-ALPNet, and training as described in Section \ref{sec:method}, we can see a boost in the segmentation results. The DINOv2 based model surpasses the Dice score of the original SSL-ALPNet across all modalities, and comes close to and even achieves better performance than SOTA models in MRI for RK and in CT for Spleen and Liver. It is also important to note that it surpasses all SOTA models in the Mean score of the tasks as seen in the last column of Table~\ref{table:mri} and Table~\ref{table:ct}.
Using a connected-components analysis (CCA) which requires no further training, increases the model's results.
We observe that incorporating TTT can slightly enhance our results compared to solely using CCA, maintaining comparable performance. This approach achieved the highest mean score on the CT dataset and the second-highest on the MRI dataset.

\noindent \textbf{Ablation Study.}
One may inquire whether there are better options to use DINOv2 for FSS. Here we compare our solution with ALPNet to other possible strategies. The first is a straightforward approach of fine-tuning the model on the support set before testing, where a simple linear layer is used as the segmentation head, as described in \cite{dinov2}. Another option is combining the  DINOv2 encoder with Mask2Former \cite{mask2former}, which a powerful segmentation model. We also compare to vanilla Mask2Former adapted to our data. The CT dataset was used for the comparisons using two experiments.
In the first, each model underwent initial supervised pre-training on a specific organ set, using the same setting described in Section \ref{sec:experiments}. Subsequently, we fine-tuned the models with three examples from an unseen organ set (the support set) and evaluated their performance on the new organ set.
In the second experiment, we bypassed the supervised pre-training step and directly fine-tuned the pre-trained models on the support set. We then assessed the model's performance on that organ set. Comparing the results of these methods in Table~\ref{table:ablation ct} to the ones of our solution with ALPNet in Table \ref{table:ct} clearly shows the advantage of our proposed solution over the other possible DINOv2 based solutions that we tested. 

\begin{table}[ht]
\centering
\resizebox{\linewidth}{!}{
\begin{tabular}{lcccccc}
\toprule
\multirow{2}{*}{Method} & \multicolumn{2}{c}{Abdominal-CT Lower} & \multicolumn{2}{c}{Abdominal-CT Upper} \\
& LK & RK & Spleen & Liver & \textbf{Mean} \\
%\cmidrule(lr){1-1}\cmidrule(lr){2-3}\cmidrule(lr){4-5}\cmidrule(lr){6-6}
\midrule
\multicolumn{6}{c}{Supervised} \\
\midrule
Mask2Former & 36.7 & 48 & 71.04 & 83.09 & 59.71 \\
Mask2Former + DINOv2 encoder  & 34.43 & 64.39 & 79.2 & 86.49 & 66.13 \\
DINOv2 encoder & 64.72 & 66.81 & 71.22 & 73.57 & 69.08 \\
\midrule
\multicolumn{6}{c}{Few Shot Results Before Pre-Training} \\
\midrule
Mask2Former & 37.07 & 28.58 & 68.37 & 73.18 & 51.8 \\
Mask2Former + DINOv2 encoder  & 15.21 & 26.15 & 62.86 & 70.95 & 47.79 \\
DINOv2 encoder & 47.96 & 48.10 & 55.02 & 66.23 & 54.32 \\
\midrule
\multicolumn{6}{c}{Few Shot Results After Pre-Training} \\
\midrule
Mask2Former & 24.09 & 20.12 & 68.73 & 71.48 & 46.10 \\
Mask2Former + DINOv2 encoder  & 10.32 & 51.95 & 74.27 & 75.67 & 53.05 \\
DINOv2 encoder & 38.89 & 37.91 & 62.02 & 66.86 & 51.42 \\
\bottomrule
\end{tabular}
}
\vspace{-0.15in}
\caption{\textbf{Ablation.} Evaluating different DINOv2 solutions.} % TODO find better caption
\label{table:ablation ct}
\vspace{-0.3in}
\end{table}

\section{Conclusion}
This paper demonstrates how a strong self-supervised model such as DINOv2 can improve semantic segmentation. We replace the ALPnet encoder with the DINOv2 ViT-based model \cite{dinov2,vit}, and show its effectiveness after fine-tuning for medical image segmentation. Our approach consistently ranked highest in the mean outcomes across the tasks.
Our results demonstrate the efficacy of this approach in handling the challenges posed by limited labeled data, making it a promising avenue for advancing the field of medical image segmentation. 

%It is important to note that CRAPNet \cite{crapnet}, outperformed our approach, particularly in the MRI scenario, across tasks such as Spleen, Liver, and Left Kidney. 
%We further improved our results through straightforward techniques, such as using Connected Component Analysis to select the most confident pixels for predictions. We also employed a simple metric for measuring confidence.
%Additionally, we found that employing a basic linear layer to process three consecutive slices before feeding them to the encoder enhanced our results, on the CT dataset. In the future we plan to find new ways to make our results better without having to train the model extensively and improving how we use pseudo-labels.

\clearpage
\section{Compliance with ethical standards}
\label{sec:ethics}
This research study was conducted retrospectively using
human subject data made available in open access by (Source
information). Ethical approval was not required as confirmed by
the license attached with the open access data.

\section{Acknowledgments}
The research in this publication was supported in part by the Israel Science Foundation (ISF) grant number 20/2629, the Israel Ministry of Science and Technology, and KLA research fund.

\bibliographystyle{IEEEbib}
\bibliography{refs}

\end{document}